\documentclass{article}

\usepackage{arxiv}

\usepackage[utf8]{inputenc} 
\usepackage[T1]{fontenc}    
\usepackage{hyperref}       
\usepackage{url}            
\usepackage{booktabs}       
\usepackage{amsfonts}       
\usepackage{nicefrac}       
\usepackage{microtype}      
\usepackage{lipsum}
\usepackage{graphicx}
\graphicspath{ {./images/} }
\usepackage{natbib}
\usepackage{booktabs}
\usepackage[table]{xcolor}
\usepackage{graphicx}
\usepackage{amsmath}
\usepackage{amssymb}
\usepackage{multirow}
\usepackage{array}
\usepackage{tabularx}
\usepackage{enumitem}
\usepackage{float}
\usepackage{placeins}

\setlist[itemize]{leftmargin=*,topsep=2pt,itemsep=1pt}
\newcolumntype{Y}{>{\raggedright\arraybackslash}X}

\definecolor{TableRule}{RGB}{100,100,100}
\definecolor{GroupGeneral}{RGB}{240,244,249}
\definecolor{GroupSpatial}{RGB}{235,245,240}

\title{CST-Bench: Benchmarking Counterfactual Spatial Transition Reasoning with Causal Transition On-Policy Self-Distillation}

\author{
  Ying Li \\
  School of Artificial Intelligence and Computer Science\\
  North China University of Technology \\
  Beijing, China \\
  \texttt{ying.li@ncut.edu.cn}
  \And
  Yongle Cheng \\
  School of Artificial Intelligence and Computer Science \\
  North China University of Technology \\
  Beijing, China \\
  \texttt{chenyongle071@gmail.com}
   \And
Jing Li \\
School of Artificial Intelligence \\
Jilin University \\
Changchun, China \\
\texttt{lij25@mails.jlu.edu.cn}
    \And
  Xingquan Cai \\
  School of Artificial Intelligence and Computer Science \\
  North China University of Technology \\
  Beijing, China \\
  \texttt{caixingquan@ncut.edu.cn}
  \And
  Yawei Luo \\
ReLER Lab \\
Zhejiang University \\
Hangzhou, China \\
  \texttt{yaweiluo@zju.edu.cn}
}

\usepackage{booktabs}
\usepackage{colortbl}

\definecolor{TableRule}{HTML}{5A5A5A}
\definecolor{GroupGeneral}{HTML}{FBE5E7}
\definecolor{GroupSpatial}{HTML}{E6F3E8}
\definecolor{GroupMethod}{HTML}{F2F2F2}
\definecolor{OursRow}{HTML}{FFF7D6}

\begin{document}

\maketitle

\begin{abstract}
Spatial reasoning is a fundamental capability of multimodal large language models (MLLMs), yet existing benchmarks mainly focus on static scene understanding. When objects are moved or swapped, models must infer not only spatial relations but also how interventions change the scene state. To address this problem, we introduce CST-Bench, a benchmark for counterfactual spatial reasoning. CST-Bench contains 7,082 examples covering 6 intervention types and 12 transformation operators, with dense causal annotations including intervention sources, affected relations, expected effects, invariant relations, and metric changes. We benchmark general-purpose MLLMs on CST-Bench and further compare post-training strategies on CST-Bench and standard spatial benchmarks. Results show that methods effective for ordinary spatial reasoning can still fail under counterfactual interventions. We further formulate counterfactual spatial reasoning as causal state-transition modeling and propose CT-OPSD, which uses CST-Bench annotations to guide a teacher model in learning counterfactual state transitions and causal reasoning chains. These chains are then distilled into a student model that reasons only from real observations. Overall, experiment results establish CST-Bench as a challenging benchmark and reusable supervision resource for counterfactual spatial reasoning. CT-OPSD further provides an initial demonstration of how its annotations can support model improvement.
\end{abstract}


\section{Introduction}

Spatial reasoning benchmarks test whether multimodal large language
models (MLLMs) understand geometric structure rather than merely
recognizing objects. Existing benchmarks cover 2D relations, depth and
distance estimation, and object orientation
~\citep{Fu2024BLINKML,Tong2024Cambrian1AF,
ma20253dsrbenchcomprehensive3dspatial}, but mainly focus on static
spatial reasoning.

However, real-world spatial reasoning also requires predicting how
relations change beyond an observed scene. Recent studies extend
spatial evaluation to large-scale urban scenes with aerial
viewpoints~\citep{Zhan2025Open3DVQAAB}, while counterfactual VQA
modifies critical image content to test whether models revise their
predictions~\citep{niu2021counterfactualvqacauseeffectlook,
chen2020counterfactualsamplessynthesizingrobust}.
SiT-Bench~\citep{Guo2026CanLS} separates spatial cognition from visual
perception, whereas
MindEdit-Bench~\citep{yu2026mindeditbenchbenchmarkingobjectlevelcounterfactual}
examines object-level counterfactual spatial reasoning in natural
images. Nevertheless, existing benchmarks either broaden scene
coverage, separate perception from reasoning, or examine isolated
counterfactual edits, providing limited supervision for complete
intervention-induced spatial transitions.

\begin{figure}[t]
\centering
\includegraphics[
  width=\columnwidth,
  pagebox=artbox
]{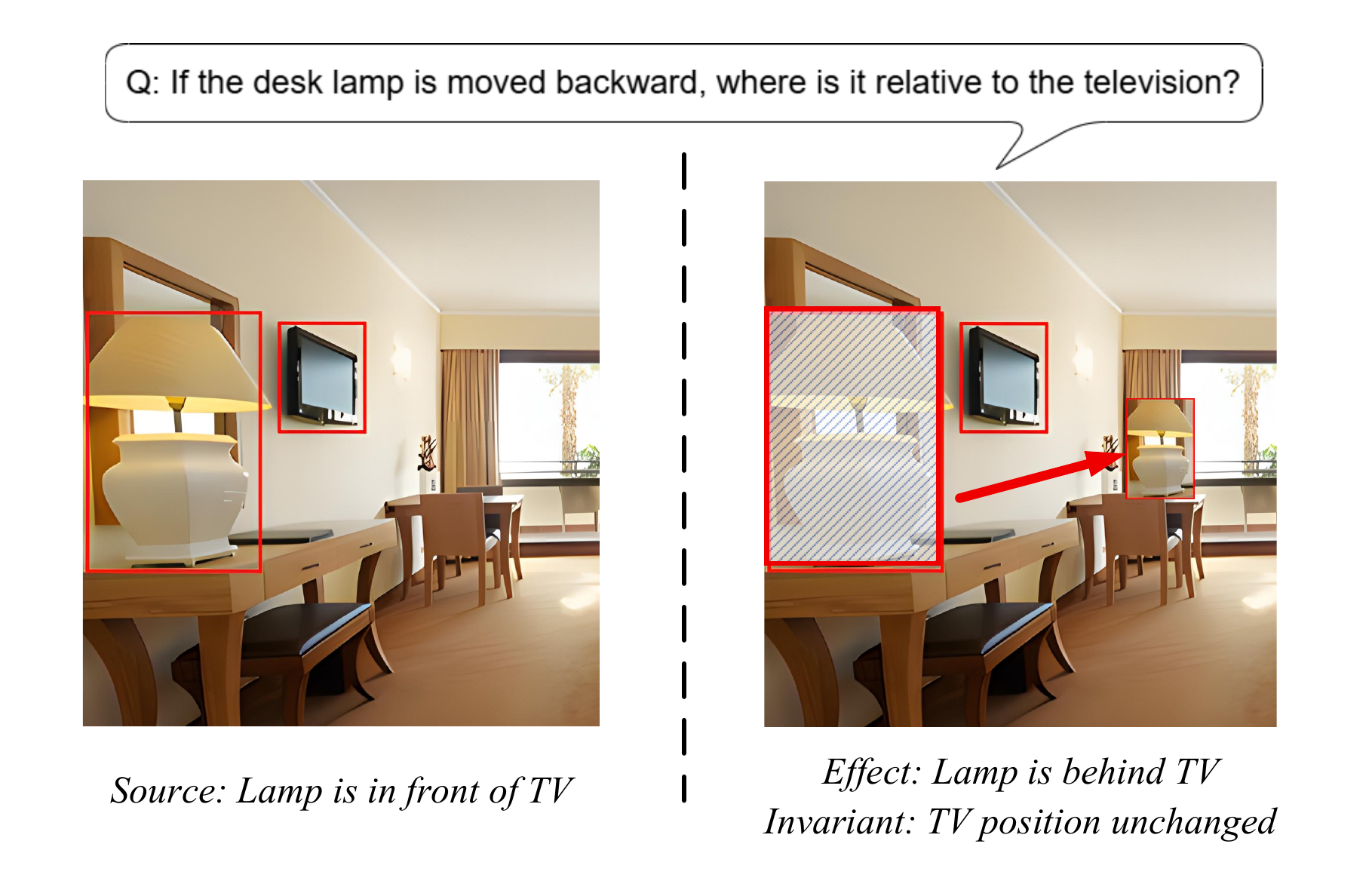}
\caption{Counterfactual spatial transition reasoning. Moving the lamp
backward changes its relation to the TV from \textit{in front of} to
\textit{behind}, while unaffected scene properties remain invariant.}
\label{fig:task_overview}
\end{figure}

We introduce CST-Bench, a benchmark centered on causal transitions
induced by spatial interventions. It contains 7,082 image-grounded
samples from distinct natural scenes, spanning 6 intervention
families and 12 fine-grained transition operators. Each sample
pairs a single query and target answer with dense annotations of source
and target states, affected relations, expected effects, preserved
invariants, and applicable metric changes. Final-answer accuracy is the
primary metric, while these annotations support diagnostic analysis and
structured causal-transition supervision. Figure~\ref{fig:task_overview}
illustrates one complete transition from an observed state to its
counterfactual state.

We benchmark general-purpose MLLMs on CST-Bench and compare several
post-training strategies using a common backbone. Results reveal that
existing models often fail to update affected relations while
preserving invariants. Because this task requires tracing the
consequences of an intervention, we formulate it as causal
state-transition modeling
~\citep{Urpi2024CausalAI,Peters2017ElementsOC,
pearl2009causality} and introduce CT-OPSD, a benchmark-aligned causal
transition on-policy self-distillation method. Distinct from existing
OPSD approaches, CT-OPSD explicitly trains its teacher with structured
source states, intervention effects, invariants, and target states,
without exposing these privileged annotations to the student.
On-policy alignment instead transfers the teacher's causal-transition
behavior to a student that uses only the image and question. Overall,
CST-Bench provides a fine-grained evaluation and supervision resource,
while CT-OPSD demonstrates how its annotations can improve
counterfactual spatial reasoning while retaining strong performance on
standard spatial benchmarks.

Our main contributions are summarized as follows:
\begin{itemize}
    \item We formulate counterfactual spatial transition reasoning as a
    standalone evaluation task requiring MLLMs to predict
    intervention-induced relational changes and preserved spatial
    invariants.

    \item We introduce CST-Bench, which grounds six interven-
tion families and twelve transition operators in 7,082 dis-
tinct natural scenes, providing systematic coverage of both
changed and invariant spatial relations.

    \item We benchmark general-purpose MLLMs and systematically compare
    post-training strategies, revealing a significant gap between
    static spatial understanding and counterfactual transition
    reasoning.

    \item We propose CT-OPSD, a benchmark-aligned on-policy
    self-distillation method that leverages privileged causal-transition
    supervision to improve counterfactual spatial reasoning while
    preserving the resulting student model's general spatial capabilities.
\end{itemize}

\section{Related Work}

\subsection{Spatial Reasoning Benchmarks}
Spatial reasoning benchmarks have expanded beyond traditional visual
question answering. Large-scale spatial VQA supervision has been used
for qualitative relations and metric estimation
\citep{chen2024spatialvlmendowingvisionlanguagemodels}, while
multifaceted reasoning has been evaluated under common and uncommon
camera viewpoints
\citep{ma20253dsrbenchcomprehensive3dspatial}. Evaluations now cover
large-scale scenes \citep{Zhan2025Open3DVQAAB} and object, camera, and
regional relations across multiple images
\citep{yang2026mmsibenchbenchmarkmultiimagespatial}. Recent work also
unifies diverse spatial tasks within broader evaluation frameworks
\citep{wu2026spatialscorecomprehensiveevaluationspatial,
cai2025holisticevaluationmultimodalllms}.

Collectively, these benchmarks broaden spatial evaluation from pairwise
relations to scene-level and multi-view reasoning, but mainly evaluate
relations within a fixed observed state rather than how that state
changes under an intervention. Their results also show that MLLMs
perform unevenly across spatial domains, viewpoints, modalities, and
reasoning skills. Spatially enhanced supervision may improve specific
tasks, but does not guarantee a consistent understanding of relational
changes. CST-Bench complements these benchmarks by testing whether
models update affected relations after an intervention while preserving
unaffected relations.

\subsection{Counterfactual Spatial Reasoning}
Spatial reasoning benchmarks have expanded from situated reasoning under
imagined egocentric perspectives
\citep{ma2023sqa3dsituatedquestionanswering}, ground-truth 3D
annotations across indoor, outdoor, and simulated scenes
\citep{cheng2024spatialrgptgroundedspatialreasoning}, and camera- and
human-centered localization
\citep{li2025viewspatialbenchevaluatingmultiperspectivespatial}, to
video-based observation
\citep{yang2025thinkingspacemultimodallarge} and large-scale 3D
spatial QA derived from raw video
\citep{gao2026holispatialevolvingvideostreams}. However, these
benchmarks primarily recover relations from observed or reconstructed
scenes while keeping the underlying scene state unchanged. SiT-Bench
separates spatial reasoning from visual perception
\citep{Guo2026CanLS}, while MindCube evaluates occluded objects and
unseen space from limited multi-view observations
\citep{wang2026mindcubespatialmentalmodeling}. These settings assess
world modeling and reference-frame transformation, but mainly alter the
observer rather than the scene state.

Counterfactual reasoning has also been explored in visual QA and
event-centric video understanding
\citep{niu2021counterfactualvqacauseeffectlook,
chen2020counterfactualsamplessynthesizingrobust,
yi2020clevrercollisioneventsvideo}. MindEdit-Bench further evaluates
object-level counterfactual editing
\citep{yu2026mindeditbenchbenchmarkingobjectlevelcounterfactual}.
Nevertheless, existing benchmarks provide limited supervision for
diagnosing the complete causal transition induced by an intervention.
CST-Bench addresses this gap by formulating counterfactual spatial
reasoning as an intervention-induced state transition with dense causal
annotations, supporting both final-answer evaluation and diagnosis of
transition failures.

\subsection{On-Policy Self-Distillation}
Traditional knowledge distillation typically relies on fixed teacher
trajectories
\citep{hinton2015distillingknowledgeneuralnetwork,
kim2016sequencelevelknowledgedistillation,bengio2015scheduled}, which
can create a mismatch between training and student inference.
Generalized distillation connects knowledge distillation with learning
using privileged information
\citep{lopezpaz2016unifyingdistillationprivilegedinformation,
vapnik2009new}. More recently, on-policy distillation aligns teacher
and student over student-generated trajectories
\citep{agarwal2024onpolicydistillationlanguagemodels}, including
teacher and student policies constructed from the same model but with
different contexts
\citep{zhao2026selfdistilledreasoneronpolicyselfdistillation}. This
paradigm has also been extended to multimodal tasks. Vision-OPD uses
local crops as privileged teacher inputs and transfers fine-grained
perception to a student that observes only the full image
\citep{yuan2026visionopdlearningfinedetails}. However, privileged
conditions may create advantages that the student cannot learn,
resulting in the ``privilege illusion''
\citep{yu2026dopddualonpolicydistillation}. CT-OPSD instead uses
CST-Bench annotations to train a causal teacher and distills its
transition reasoning into a student policy; the annotations are used
only during training, while the student relies solely on the image and
question at inference time.

\begin{figure*}[t]
\centering
\includegraphics[width=\textwidth]{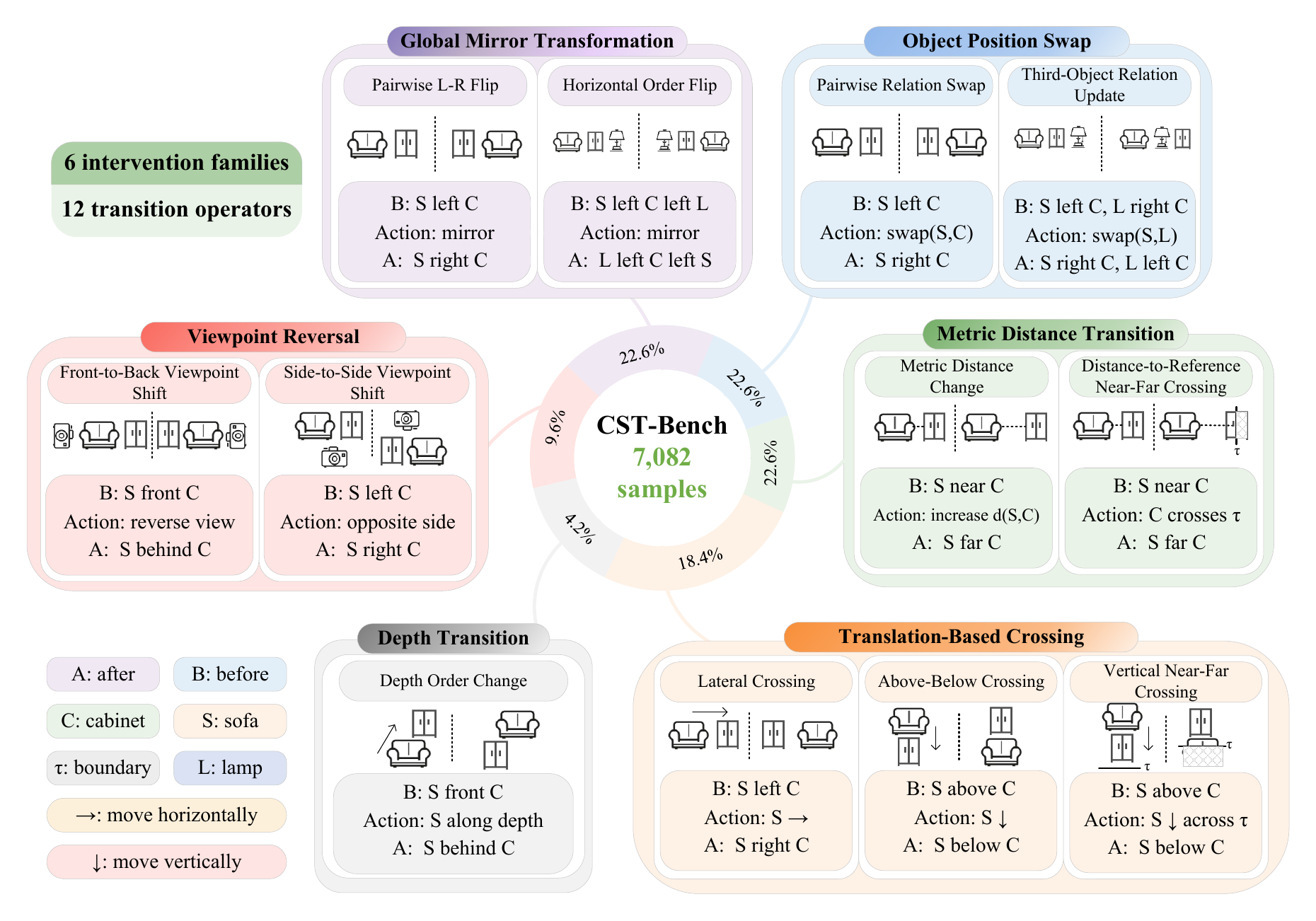}
\caption{Overview of the CST-Bench transition taxonomy.
CST-Bench contains 7,082 counterfactual spatial QA samples across
six intervention families and twelve transition operators. The
operators cover relational state updates, metric distance changes, and
qualitative relation-threshold crossings, with dense causal transition
annotations.}
\label{fig:cst_taxonomy}
\end{figure*}

\section{CST-Bench: A Counterfactual Spatial Reasoning Benchmark}
\label{sec:cst_bench}
CST-Bench frames counterfactual spatial reasoning as updating a source
state after an explicit intervention. It contains six intervention
families and twelve fine-grained operators
(Figure~\ref{fig:cst_taxonomy}), constructed through deterministic
generation, VLM consensus validation, and human verification. Its dense
annotations support diagnostic evaluation and enable CT-OPSD, an
auxiliary approach that transfers transition reasoning to an
observation-only student.

\subsection{Task Taxonomy}

CST-Bench organizes twelve transition operators into six intervention
families covering global, local, metric, translation, depth, and
viewpoint changes. Across 7,082 image-grounded samples, it recasts
static observations as intervention-conditioned transitions, jointly
testing whether models update affected relations and preserve
invariants.

\paragraph{Global Mirror Transformation.}
This family tests globally consistent horizontal updates
\citep{ma20253dsrbenchcomprehensive3dspatial}. Pairwise Left--Right
Flip reverses a single relation, whereas Multi-object Horizontal Order
Flip updates multiple objects consistently. Both preserve object
identities, object count, and vertical relations.

\paragraph{Object Position Swap.}
This family applies identity-preserving local interventions
\citep{yu2026mindeditbenchbenchmarkingobjectlevelcounterfactual}.
Pairwise Relation Swap tests the direct effect between exchanged
objects, while Third-object Relation Update requires propagating the
change to an unchanged reference object.

\begin{figure*}[t]
\centering
\includegraphics[width=\textwidth]{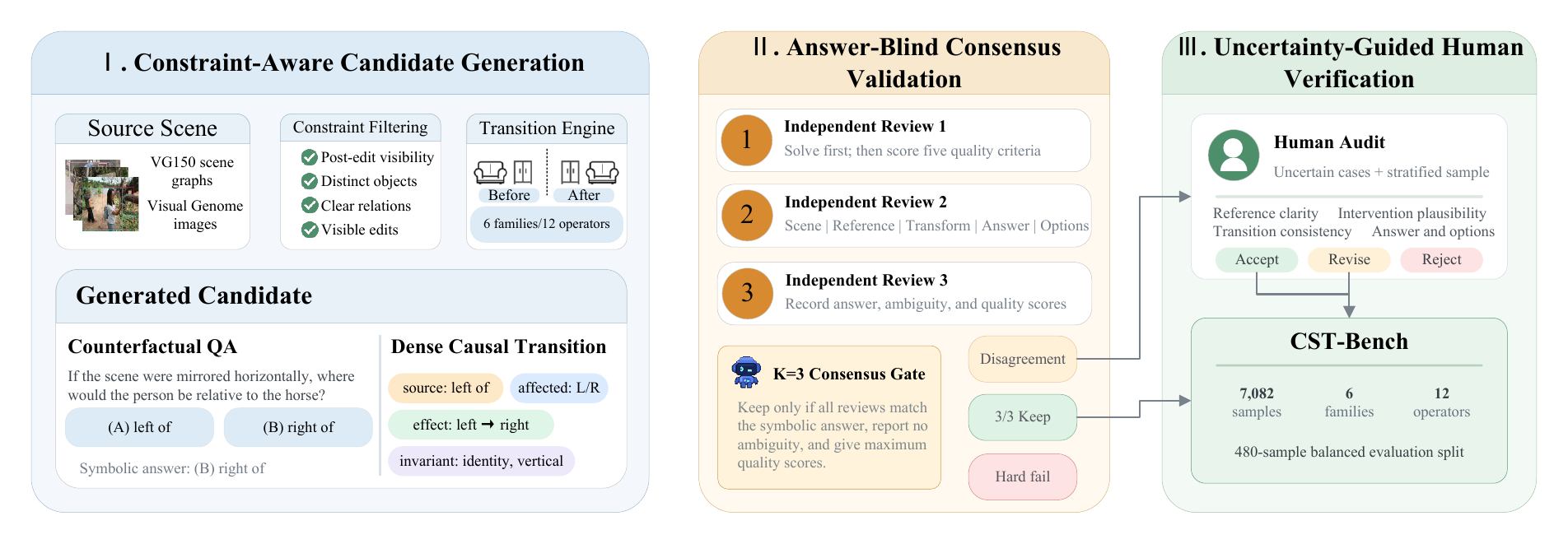}
\caption{Overview of CST-Bench construction. Constraint-aware
generation jointly derives counterfactual states and annotations,
answer-blind VLM consensus filters ambiguous candidates, and
uncertainty-guided human verification resolves uncertain cases and
audits all transformation operators.}
\label{fig:construction_pipeline}
\end{figure*}

\paragraph{Metric Distance Transition.}
This family distinguishes continuous distance sensitivity from
categorical relation updates
\citep{chen2024spatialvlmendowingvisionlanguagemodels,
ma20253dsrbenchcomprehensive3dspatial}. Metric Distance Change
(Relation Preserved) changes distance while retaining the qualitative
relation, whereas Distance-to-Reference Near--Far Crossing moves it
across the \emph{near}--\emph{far} threshold
\citep{cohn2008qualitative}.

\paragraph{Translation-Based Crossing.}
This family uses object translation to induce spatial relation changes
\citep{yu2026mindeditbenchbenchmarkingobjectlevelcounterfactual}.
Lateral Relation Crossing and Above--Below Crossing test horizontal and
vertical updates, while Vertical Near--Far Crossing changes the
near--far relation through vertical displacement. Unaffected relations
must remain unchanged.

\paragraph{Depth Transition.}
Depth Order Change moves an object across a reference along the depth
axis, extending observed front--behind evaluation to intervention-based
state updates
\citep{Fu2024BLINKML,ma20253dsrbenchcomprehensive3dspatial}. The depth
ordering changes while image-plane relations are preserved.

\paragraph{Viewpoint Reversal.}
This family changes the observer while preserving the physical scene.
Front-to-Back and Side-to-Side Viewpoint Shifts require recomputing
observer-centered relations without altering the allocentric layout
\citep{li2025viewspatialbenchevaluatingmultiperspectivespatial,
wang2026mindcubespatialmentalmodeling,klatzky1998allocentric}.

\subsection{Benchmark Construction}

As shown in Fig.~\ref{fig:construction_pipeline}, unlike standard
spatial QA, the post-intervention state is not directly observed and
must be inferred from a hypothetical
intervention~\citep{Zhang_2024_CVPR}. A valid sample must realize the
intended effect, propagate dependent relation changes, preserve
invariants, and remain visually unambiguous. We address this challenge
through three stages.

\begin{figure*}[t]
    \centering
    \includegraphics[width=\textwidth]{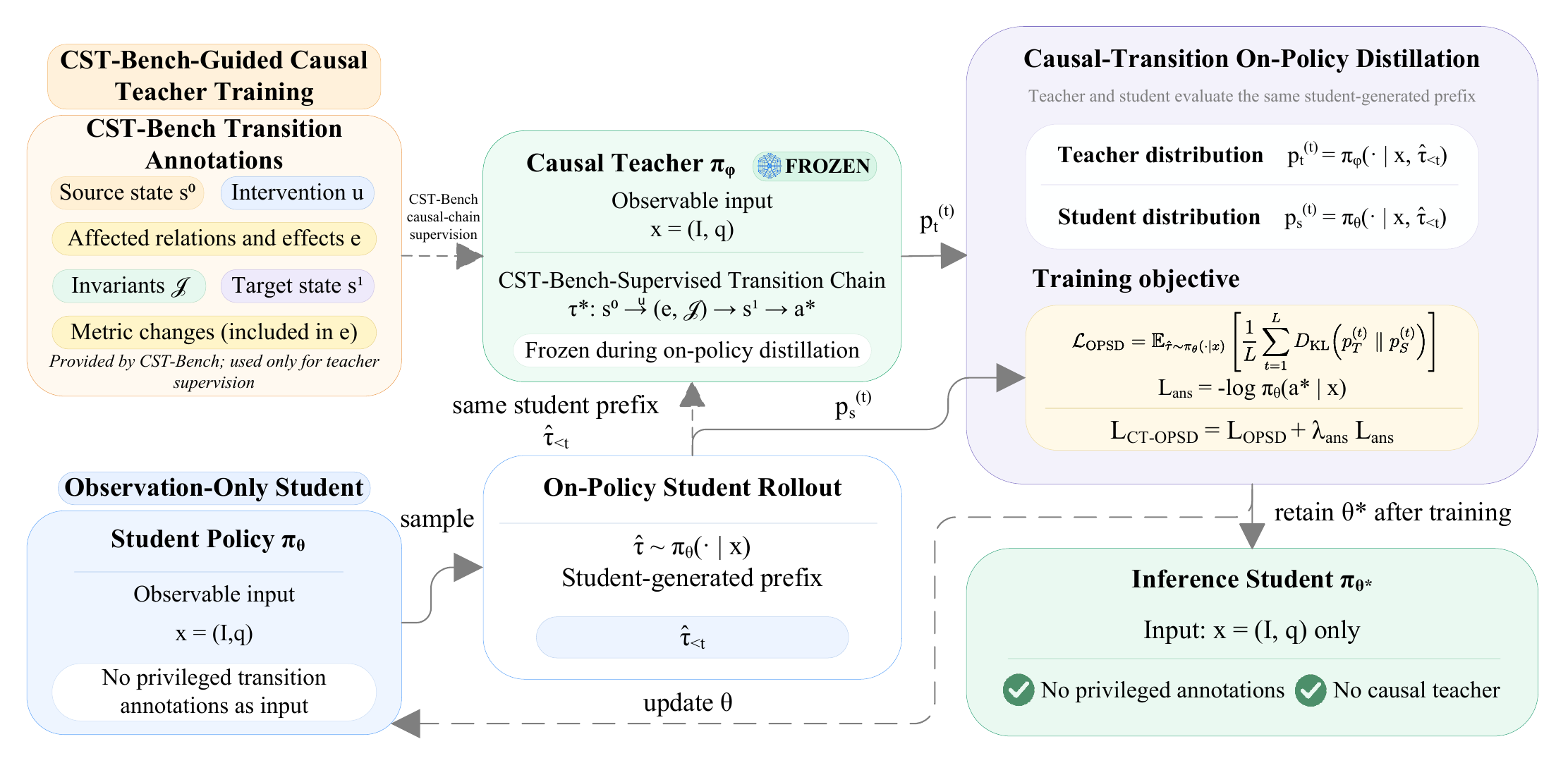}
    \caption{Overview of CT-OPSD with CST-Bench supervision. Privileged
    transition annotations from CST-Bench train a causal teacher but are
    never provided as student inputs. The frozen teacher aligns with the
    observation-only student on student-generated prefixes, and only the
    updated student is retained for inference.}
    \label{fig:ct_opsd}
\end{figure*}

\paragraph{Stage I: Constraint-Aware Candidate Generation.}
We curate scenes from the VG150 split of Visual
Genome~\citep{xu2017scenegraphgenerationiterative,
krishna2016visualgenomeconnectinglanguage} with valid object boxes,
sufficient distinct objects, and geometrically unambiguous relations.
Pairwise operators require two usable objects, whereas Third-object
Relation Update requires three objects from distinct categories.
Relations are computed from normalized bounding boxes, with positional
references used when necessary.

Prior work shows that automatically generated spatial QA may contain
ambiguous references and shortcut-prone or unreliable
labels~\citep{chen2024spatialvlmendowingvisionlanguagemodels,
ma20253dsrbenchcomprehensive3dspatial}. To address these risks, a
deterministic, operator-aware editor applies each intervention, and a
symbolic solver jointly derives the pre- and post-intervention states,
counterfactual QA pair, affected relations, expected effects,
invariants, and applicable metric changes. Deriving these fields from
the same edited state prevents contradictions between the transition
and its dense annotations. We reject candidates that violate
operator-specific relation constraints, contain invalid boxes or
ambiguous relations, provide insufficient relation separation, or move
objects outside the image boundary.

We score each remaining candidate by
\begin{equation}
s(c)=s_{\mathrm{ref}}+s_{\mathrm{rel}}+s_{\mathrm{edit}}
+s_{\mathrm{vis}}+s_{\mathrm{nat}},
\end{equation}
where the terms measure reference clarity, relation clarity, edit
effectiveness, post-edit visibility, and transformation naturalness.
Candidates with $s(c)<8$ are removed. The remaining candidates are
ranked by
\begin{equation}
r(c)=\frac{1}{2}\hat{s}_{\mathrm{op}}(c)
+\frac{1}{2}\hat{s}_{\mathrm{global}}(c),
\end{equation}
where the normalized scores are computed within an operator and
globally, respectively. We retain at most one highest-ranked
transformation per source image.

\paragraph{Stage II: Answer-Blind Consensus Validation.}
Although LLM/VLM-based judging provides scalable validation, prior work
has identified systematic biases and limited reliability in
single-pass evaluation
~\citep{zheng2023judgingllmasajudgemtbenchchatbot,
wang2023largelanguagemodelsfair}. Drawing on multi-sample
consistency~\citep{wang2023selfconsistencyimproveschainthought}, each
candidate is independently reviewed $K=3$ times by a VLM. To prevent
the symbolic label from influencing the initial solution, the VLM first
solves the problem without accessing the generated answer and only then
compares its prediction with the symbolic label. It also scores
scene--image consistency, reference clarity, transformation fidelity,
answer correctness, and option quality from 0 to 2. A candidate passes
only under unanimous agreement:
\begin{equation}
\operatorname{Keep}(c)=
\bigwedge_{k=1}^{K}
\left[
\hat{a}_{k}=a^{*}
\land b_{k}=0
\land \min_j q_{k,j}=2
\right],
\end{equation}
where $\hat{a}_{k}$ is the independently predicted answer, $a^{*}$ is
the symbolic answer, $b_k$ indicates ambiguity, and $q_{k,j}$ is the
score for criterion $j$. Candidates with repeated hard failures are
removed, while unresolved cases are routed to human verification.

\paragraph{Stage III: Uncertainty-Guided Human Verification.}
Human reviewers inspect unresolved candidates and a stratified sample
covering all transformation operators. They verify reference clarity,
intervention plausibility, transition consistency, answer correctness,
and option quality. Ambiguous or inconsistent samples are corrected
when possible and otherwise excluded.

\section{Causal Transition On-Policy Self-Distillation}

As an application of CST-Bench's dense annotations, CT-OPSD models
counterfactual spatial reasoning as causal state-transition prediction.
Distinct from standard OPSD, it explicitly trains a teacher to organize
source states, intervention effects, invariants, and target states into
a structured transition chain. The resulting behavior is distilled
through on-policy distribution matching, while the privileged
annotations are never provided as student inputs
~\citep{agarwal2024onpolicydistillationlanguagemodels,
zhao2026selfdistilledreasoneronpolicyselfdistillation}.

\paragraph{Causally Structured Teacher.}
Let $x=(I,q)$ denote the observable input, where $I$ is the image and
$q$ contains the question, intervention description, and answer
options. CST-Bench additionally provides privileged transition
annotations
\begin{equation}
z_c=(s^0,u,e,\mathcal{I},s^1),
\end{equation}
where $s^0$ and $s^1$ are the source and target states, $u$ is the
structured intervention, $e$ records affected relations and expected
effects, and $\mathcal{I}$ contains preserved invariants. When
applicable, $e$ also includes the corresponding metric change. Together
with the standard answer label $a^*$, these fields form
\begin{equation}
\tau^*: s^0 \xrightarrow{\,u\,} (e,\mathcal{I})
\xrightarrow{} s^1 \xrightarrow{} a^*.
\end{equation}
The teacher $\pi_\phi$ learns to generate this chain from $x$:
\begin{equation}
\mathcal{L}_{\mathrm{teacher}}
=
-\sum_{t=1}^{|\tau^*|}
\log \pi_\phi(\tau_t^* \mid x,\tau_{<t}^*).
\end{equation}
The teacher is frozen after this stage.

\begin{table*}[t]
\centering
\caption{Performance of MLLMs on CST-Bench (\%) under the answer-only
protocol on the balanced 480-example evaluation split. Best and
second-best results are bolded and underlined, respectively.}
\label{tab:foundation_models}
\small
\setlength{\tabcolsep}{5pt}
\renewcommand{\arraystretch}{1.10}
\arrayrulecolor{TableRule}
\begin{tabular}{l|cccccc|cc}
\toprule
& \multicolumn{6}{c|}{\textbf{Transition Families}} &
\multicolumn{2}{c}{\textbf{Aggregate Results}} \\
\textbf{Model} & \textbf{Mirror} & \textbf{Swap} & \textbf{Metric} &
\textbf{Translation} & \textbf{Depth} & \textbf{Viewpoint} &
\textbf{Macro Avg.} & \textbf{Micro Acc.} \\
\midrule
\rowcolor{GroupGeneral}
\multicolumn{9}{c}{\emph{\textbf{General-purpose Open-source MLLMs}}} \\
Qwen2.5-VL-3B
& 15.5 & 15.1 & 17.8 & 20.7 & 11.6 & 17.0 & 16.3 & 17.0 \\
Qwen3-VL-4B
& 42.5 & 25.0 & 35.6 & \textbf{96.3} & \textbf{37.5}
& 30.0 & \underline{44.5} & \textbf{49.4} \\
Qwen2.5-VL-7B
& 27.6 & 19.2 & 22.3 & 36.3 & 12.5 & 20.6 & 23.1 & 25.1 \\
LLaVA-OneVision-7B
& 26.3 & 35.0 & 39.4 & \underline{90.0} & 32.5
& 27.5 & 41.8 & 46.6 \\
InternVL3.5-4B
& 32.5 & 25.0 & 26.2 & 78.3 & 32.5 & 30.0 & 37.4 & 41.2 \\
Phi-3.5-Vision-4B
& 33.8 & 22.5 & 33.8 & 85.0 & 17.5 & 31.2 & 37.3 & 42.9 \\
Gemma-3-4B-it
& \textbf{52.5} & \textbf{42.5} & \underline{42.5} & 60.0
& \underline{35.0} & \textbf{40.0} & \textbf{45.4}
& \underline{47.5} \\
\midrule
\rowcolor{GroupSpatial}
\multicolumn{9}{c}{\emph{\textbf{Spatial- and Embodied-oriented Open-source MLLMs}}} \\
SpatialThinker-3B
& 27.5 & 31.2 & 37.5 & 70.0 & 22.5 & 26.2 & 35.8 & 39.8 \\
RoboBrain2.0-7B
& \underline{50.0} & \underline{37.5} & 21.3 & 33.8
& 5.0 & 27.5 & 29.2 & 31.6 \\
SpaceR-SFT-7B
& 41.3 & 20.0 & \textbf{51.3} & 70.8 & 32.5
& \underline{32.5} & 41.4 & 44.6 \\
\bottomrule
\end{tabular}
\arrayrulecolor{black}
\end{table*}

\paragraph{Student Rollout and Causal Bridge.}
The student $\pi_\theta$ is initialized from the same base model and
uses only $x$ as input. It generates an on-policy trajectory
\begin{equation}
\hat{\tau}\sim\pi_\theta(\cdot\mid x)
\end{equation}
~\citep{ross2011reductionimitationlearningstructured}. At each step,
the frozen teacher and student evaluate the same student-generated
prefix:
\begin{equation}
p_T^{(t)}=\pi_\phi(\cdot\mid x,\hat{\tau}_{<t}),\qquad
p_S^{(t)}=\pi_\theta(\cdot\mid x,\hat{\tau}_{<t}).
\end{equation}
Thus, the causal bridge transfers the teacher's structured transition
behavior on states encountered by the student, without exposing
$z_c$ or $\tau^*$ to the student.

\paragraph{Training Objective.}
We align the two distributions along student rollouts using forward KL
divergence:
\begin{equation}
\mathcal{L}_{\mathrm{OPSD}}
=
\mathbb{E}_{\hat{\tau}\sim\pi_\theta(\cdot\mid x)}
\left[
\frac{1}{L}\sum_{t=1}^{L}
D_{\mathrm{KL}}\!\left(
p_T^{(t)}\parallel p_S^{(t)}
\right)
\right].
\end{equation}
The complete objective additionally includes standard answer
supervision:
\begin{equation}
\mathcal{L}_{\mathrm{CT\text{-}OPSD}}
=
\mathcal{L}_{\mathrm{OPSD}}
+\lambda_{\mathrm{ans}}
\left[-\log\pi_\theta(a^*\mid x)\right].
\end{equation}
At inference, the teacher and all privileged transition annotations are
discarded, and the student predicts from $x=(I,q)$ alone.

\section{Experiments}

\subsection{Implementation Details}
We evaluate ten open-source multimodal large language models (MLLMs):
Qwen2.5-VL-3B and Qwen2.5-VL-7B, Qwen3-VL-4B
\citep{bai2025qwen25vltechnicalreport}, LLaVA-OneVision-7B
\citep{li2024llavaonevisioneasyvisualtask}, InternVL3.5-4B
\citep{wang2025internvl35advancingopensourcemultimodal},
Phi-3.5-Vision-4B \citep{abdin2024phi3technicalreporthighly},
Gemma-3-4B-it \citep{gemmateam2025gemma3technicalreport},
RoboBrain2.0-7B \citep{baairobobrainteam2025robobrain20technicalreport},
SpatialThinker-3B
\citep{batra2026spatialthinkerreinforcingscenegraphgrounded}, and
SpaceR-SFT-7B \citep{ouyang2025spacerreinforcingmllmsvideo}.
All foundation models are evaluated without fine-tuning on CST-Bench.

We use the balanced 480-example evaluation split, which contains 40
examples for each of the twelve fine-grained operators. These operators
are grouped into six transition families: Global Mirror Transformation,
Object Position Swap, Metric Distance Transition, Translation-Based
Crossing, Depth Transition, and Viewpoint Reversal. All retained
examples keep the referenced objects visible before and after the
intervention; distance-related operators test continuous metric changes
or qualitative near--far threshold crossings rather than visibility
changes.

At inference time, every model receives only the image, the
counterfactual question, and the answer options. Models do not access
scene states, intervention effects, invariant relations, or other
privileged causal annotations. We use a common answer-only prompt,
greedy decoding, and a maximum of 64 newly generated tokens. Each
response is parsed as a multiple-choice answer.

\subsection{Baselines and Evaluation Metrics}
Using Qwen2.5-VL-3B as the common base, we compare the unadapted
model, answer-supervised SFT, GRPO
\citep{shao2024deepseekmathpushinglimitsmathematical}, OPSD
\citep{zhao2026selfdistilledreasoneronpolicyselfdistillation}, and
CT-OPSD. All variants use the same observable input $(I,q)$; privileged
annotations are available only when training the CT-OPSD teacher.

We report multiple-choice accuracy. Foundation models are evaluated by
family accuracy, Macro Avg.\ over the six families, and Micro Acc.\
over all 480 examples. Post-trained models are additionally evaluated
on 3DSRBench
\citep{ma20253dsrbenchcomprehensive3dspatial}, CV-Bench
\citep{Tong2024Cambrian1AF}, and
BLINK$_{\mathrm{val}}$ \citep{Fu2024BLINKML}.

\begin{table*}[t]
\centering
\caption{Post-training results on CST-Bench and standard spatial
benchmarks (\%). All variants use Qwen2.5-VL-3B as the base model.
Best and second-best results are bolded and underlined, respectively.}
\label{tab:training_comparison}
\small
\setlength{\tabcolsep}{4.5pt}
\renewcommand{\arraystretch}{1.12}
\arrayrulecolor{TableRule}
\begin{tabular}{l|c|c|ccc|ccc}
\toprule
& \multicolumn{1}{c|}{\textbf{CST-Bench}} &
\multicolumn{1}{c|}{\textbf{3DSRBench}} &
\multicolumn{3}{c|}{\textbf{CV-Bench}} &
\multicolumn{3}{c}{\textbf{BLINK$_{\mathrm{val}}$}} \\
\textbf{Model} &
\textbf{Micro Acc.} & \textbf{Acc.} &
\textbf{2D} & \textbf{3D} & \textbf{Avg.} &
\shortstack{\textbf{Spatial}\\\textbf{Relation}} &
\shortstack{\textbf{Relative}\\\textbf{Depth}} &
\textbf{Avg.} \\
\midrule
\rowcolor{GroupMethod}
\multicolumn{9}{c}{\emph{\textbf{Post-training Strategies with Qwen2.5-VL-3B}}} \\
Qwen2.5-VL-3B
& 17.0 & 46.0 & 64.0 & 66.0 & 65.0 & \underline{83.0} & 64.0 & 73.5 \\
Qwen2.5-VL-3B + SFT
& 50.3 & 38.0 & 63.0 & 68.0 & 65.5 & 70.0 & 67.0 & 68.5 \\
Qwen2.5-VL-3B + GRPO
& 46.7 & \textbf{52.9} & 71.0 & 76.3 & 73.7 & 81.8 & 66.9 & 74.4 \\
Qwen2.5-VL-3B + OPSD
& \underline{53.3} & \underline{51.8} & \underline{73.8}
& \underline{78.3} & \underline{76.1} & 82.5
& \underline{67.2} & \underline{74.9} \\
\rowcolor{OursRow}
Qwen2.5-VL-3B + CT-OPSD
& \textbf{55.2} & 49.5 & \textbf{75.1} & \textbf{78.6}
& \textbf{76.9} & \textbf{84.0} & \textbf{69.2} & \textbf{76.6} \\
\bottomrule
\end{tabular}
\arrayrulecolor{black}
\end{table*}

For component ablation, CST W-Avg.\ denotes the predefined weighted
aggregate of causal-transition diagnostics, distinct from the
family-level Macro Avg. Its definition and weights are provided in the
appendix.

\subsection{Main Results}

Table~\ref{tab:foundation_models} compares ten open-source MLLMs on
the balanced CST-Bench evaluation split. Overall performance remains
limited: no foundation model is consistently strong across all six
transition families. Gemma-3-4B-it achieves the best Macro Avg.\ of
45.4, indicating the most balanced family-level performance, whereas
Qwen3-VL-4B obtains the highest Micro Acc.\ of 49.4.

The difference between these rankings is expected rather than
contradictory. Macro Avg.\ weights each transition family equally,
whereas Micro Acc.\ weights families according to their number of
examples. In particular, Translation-Based Crossing contains three
operators and thus contributes more to Micro Acc.\ than the
single-operator Depth Transition. We therefore report both metrics to
distinguish balanced transition reasoning from aggregate accuracy.

\subsection{Family-wise Analysis}

The leading model varies substantially by transition family. Gemma-3-4B-it
performs best on Global Mirror Transformation (52.5), Object Position
Swap (42.5), and Viewpoint Reversal (40.0). SpaceR-SFT-7B achieves the
best result on Metric Distance Transition (51.3), while Qwen3-VL-4B
leads Translation-Based Crossing (96.3) and Depth Transition (37.5).
This distribution indicates that competence in one type of spatial
update does not reliably transfer to other transformations.

Depth Transition and Viewpoint Reversal are particularly challenging:
their best scores are only 37.5 and 40.0, respectively. Qwen3-VL-4B
illustrates the resulting imbalance, achieving 96.3 on translation but
only 25.0 on swapping and 30.0 on viewpoint reversal. Thus, strong
performance on image-plane translation alone is insufficient for
reliable counterfactual spatial reasoning across relational, metric,
depth, and reference-frame changes.

\subsection{Post-training Comparison}

Table~\ref{tab:component_ablation} reports source-state understanding,
effect reasoning, invariance identification, and transition
consistency, abbreviated as Src., Eff., Inv., and Cons. Full CT-OPSD
achieves the best CST W-Avg.\ (25.9) and answer accuracy (55.2), while
all component removals reduce both metrics. Although individual
diagnostic scores vary across variants, the full method provides the
strongest overall balance.

Table~\ref{tab:training_comparison} compares post-training strategies
using Qwen2.5-VL-3B. CT-OPSD achieves the best CST-Bench Micro Acc.\
(55.2), outperforming standard OPSD by 1.9 points, and obtains the best
CV-Bench and BLINK averages. Its 3DSRBench score remains below GRPO and
OPSD but exceeds the unadapted base model, indicating improved
counterfactual reasoning without uniformly maximizing every external
benchmark.

\begin{table}[t]
\centering
\caption{Component ablation of CT-OPSD. CST W-Avg.\ is the weighted
diagnostic aggregate. Best and second-best results are bolded and
underlined, respectively.}
\label{tab:component_ablation}
\scriptsize
\setlength{\tabcolsep}{2.6pt}
\renewcommand{\arraystretch}{1.06}
\arrayrulecolor{TableRule}
\resizebox{0.5\columnwidth}{!}{%
\begin{tabular}{@{}lcccccc@{}}
\toprule
\textbf{Variant} &
\shortstack{\textbf{CST}\\\textbf{W-Avg.}} &
\shortstack{\textbf{Ans.}\\\textbf{Acc.}} &
\textbf{Src.} & \textbf{Eff.} & \textbf{Inv.} & \textbf{Cons.} \\
\midrule
OPSD
& 24.0 & \underline{53.3} & 34.1 & 0.7 & 0.3 & 22.5 \\
CT w/o Aux.
& \underline{24.7} & 52.7 & 40.1 & 0.9 & 0.4 & 30.0 \\
CT w/o Bridge
& 24.5 & 52.1 & 40.0 & 1.0 & \underline{0.7} & \underline{30.6} \\
CT w/o Eff.
& 24.0 & 50.4 & 40.0 & 1.2 & \textbf{1.0} & 28.5 \\
CT w/o Inv.
& 24.6 & 51.7 & \underline{41.2} & \underline{1.3} & 0.5 & \textbf{31.2} \\
CT w/o Metric
& 24.1 & 50.4 & \textbf{41.3} & 1.2 & 0.5 & 28.5 \\
\rowcolor{OursRow}
CT-OPSD
& \textbf{25.9} & \textbf{55.2} & \textbf{41.3} & \textbf{1.5} & 0.6 & 30.4 \\
\bottomrule
\end{tabular}%
}
\arrayrulecolor{black}
\end{table}

\section{Analysis}

The fundamental distinction of CST-Bench lies not merely in task
difficulty, but in the object of inference. Static benchmarks ask a
model to recognize a spatial state already present in an image, whereas
CST-Bench requires it to intervene on that state and predict a valid
successor under conservation constraints. It therefore tests whether
the model's spatial representation is coherent and manipulable, rather
than a collection of independently recognized relation labels. The
uneven family-level performance is consistent with current MLLMs
lacking such a unified transition model.

Dense transition annotations are important because counterfactual
correctness has two inseparable sides: applying the intended change and
avoiding unintended changes. A model may fail by ignoring the
intervention, but it may also fail by updating too much. Final-answer
supervision alone does not distinguish these errors. CT-OPSD is
effective because its supervision reflects this structure: the teacher
learns selective state updating, and on-policy alignment transfers that
behavior at states encountered by the student, without providing
privileged evidence at inference.

The ablations reinforce this interpretation. Improvements on an
isolated diagnostic dimension do not necessarily improve overall
performance because effects and invariants must be satisfied jointly.
The key connection is therefore structural: CST-Bench defines valid
spatial transitions, its annotations expose their internal constraints,
and CT-OPSD converts those constraints into a learnable training signal.

\section{Conclusion}

We introduced CST-Bench, a counterfactual spatial reasoning benchmark
with 7,082 examples across six families and twelve operators. It
evaluates intervention-induced relational, metric, depth, and viewpoint
changes with dense causal transition annotations. We also proposed
CT-OPSD, which distills a privileged causal teacher into an
observation-only student. Experiments reveal uneven counterfactual
capabilities among existing MLLMs, while CT-OPSD improves CST-Bench
accuracy and retains strong performance on standard spatial benchmarks.

\bibliographystyle{plainnat}
\bibliography{aaai2027}

\clearpage
\begin{center}
{\LARGE\bfseries Technical Supplement}
\end{center}
\appendix

This technical supplement provides additional details for CST-Bench and
CT-OPSD. It follows the task definition, construction procedure, and
observation-only inference setting described in the main paper. The main
paper is self-contained; the material below specifies the benchmark records,
operator semantics, validation protocol, and experimental diagnostics.

\section{Representative CST-Bench Examples}
\label{app:examples}

\begin{figure*}[t]
\centering
\begin{minipage}[t]{0.32\textwidth}
\centering
\includegraphics[width=\linewidth,height=0.135\textheight,keepaspectratio]{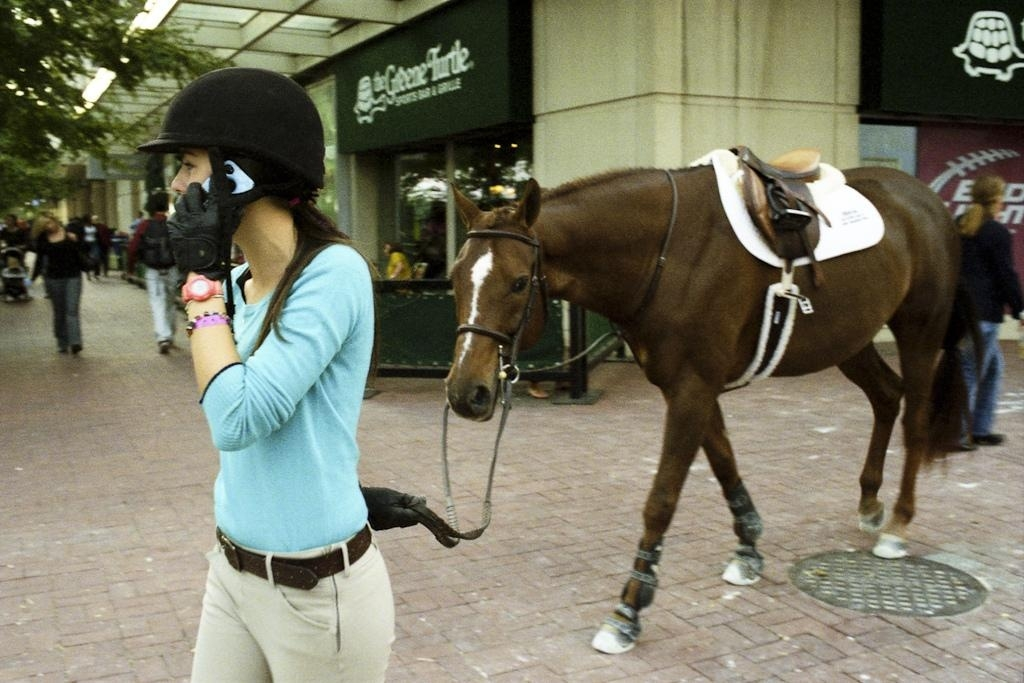}
\par\smallskip
\textbf{(a) Global mirror}
\end{minipage}\hfill
\begin{minipage}[t]{0.32\textwidth}
\centering
\includegraphics[width=\linewidth,height=0.135\textheight,keepaspectratio]{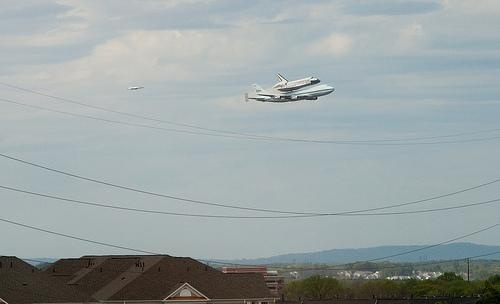}
\par\smallskip
\textbf{(b) Object swap}
\end{minipage}\hfill
\begin{minipage}[t]{0.32\textwidth}
\centering
\includegraphics[width=\linewidth,height=0.135\textheight,keepaspectratio]{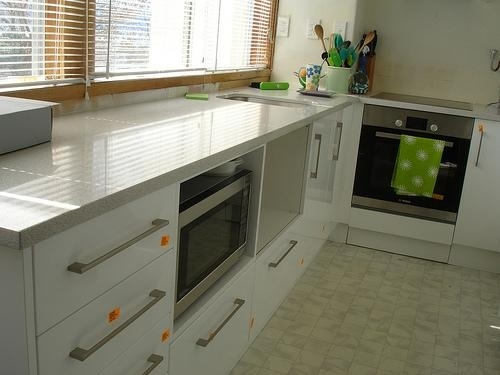}
\par\smallskip
\textbf{(c) Metric distance}
\end{minipage}

\medskip
\begin{minipage}[t]{0.32\textwidth}
\centering
\includegraphics[width=\linewidth,height=0.135\textheight,keepaspectratio]{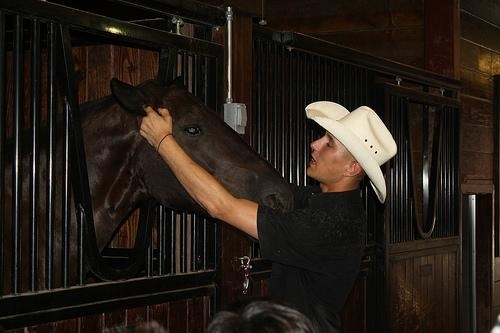}
\par\smallskip
\textbf{(d) Translation crossing}
\end{minipage}\hfill
\begin{minipage}[t]{0.32\textwidth}
\centering
\includegraphics[width=\linewidth,height=0.135\textheight,keepaspectratio]{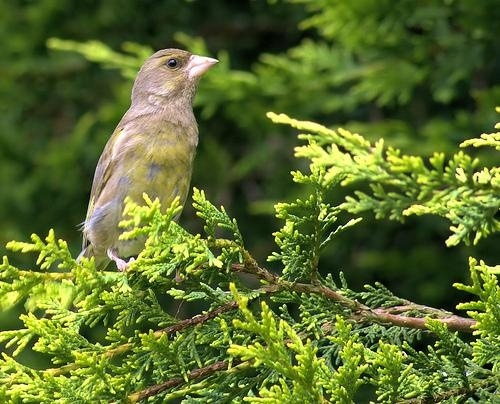}
\par\smallskip
\textbf{(e) Depth transition}
\end{minipage}\hfill
\begin{minipage}[t]{0.32\textwidth}
\centering
\includegraphics[width=\linewidth,height=0.135\textheight,keepaspectratio]{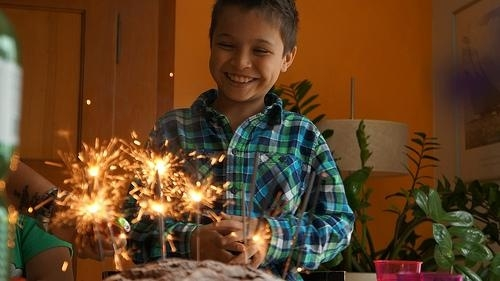}
\par\smallskip
\textbf{(f) Viewpoint reversal}
\end{minipage}
\caption{Representative source images from the six CST-Bench intervention
families. Each example is paired with a single intervention-conditioned query
and a symbolic target transition.}
\label{fig:six_examples}
\end{figure*}

Figure~\ref{fig:six_examples} presents one source image from each intervention
family. Each sample contains one observed image, one intervention-conditioned
multiple-choice question, one target answer, and one causal-transition record.
The post-intervention image is not rendered. Instead, the target state is
defined symbolically from the observed scene and the hypothetical intervention.

\paragraph{Global Mirror Transformation.}
The query asks how an object is related to a reference after a horizontal
mirror transformation. A source relation of \textit{left of} becomes
\textit{right of}; object identity, object count, and vertical relations are
preserved.

\paragraph{Object Position Swap.}
Two objects exchange positions, while the query may concern one exchanged
object and a third, unchanged reference object. The example therefore tests
whether the direct edit is propagated to the relevant third-object relation.

\paragraph{Metric Distance Transition.}
The intervention changes normalized distance while holding direction fixed.
The qualitative relation may remain unchanged or cross the \textit{near}/
\textit{far} boundary, separating metric sensitivity from categorical updates.

\paragraph{Translation-Based Crossing.}
An object is translated across a reference horizontally or vertically. The
affected relation changes, while the orthogonal relation and object properties
remain invariant.

\paragraph{Depth Transition.}
An object moves across a reference along the depth axis. The \textit{behind}/
\textit{in front of} relation changes while image-plane left--right and
above--below relations are preserved.

\paragraph{Viewpoint Reversal.}
The observer changes viewpoint while the physical scene remains fixed. The
task recomputes observer-centered relations without treating the viewpoint
change as an edit to the allocentric scene layout.

\section{CST-Bench Specification and Construction}
\label{app:cst}

\begin{table}[t]
\centering
\caption{Full-dataset distribution by intervention family.}
\label{tab:family_distribution}
\small
\begin{tabular}{lrr}
\toprule
\textbf{Family} & \textbf{Samples} & \textbf{Share} \\
\midrule
Global Mirror Transformation & 1,600 & 22.6\% \\
Object Position Swap & 1,600 & 22.6\% \\
Metric Distance Transition & 1,600 & 22.6\% \\
Translation-Based Crossing & 1,300 & 18.4\% \\
Depth Transition & 300 & 4.2\% \\
Viewpoint Reversal & 682 & 9.6\% \\
\midrule
Total & 7,082 & 100.0\% \\
\bottomrule
\end{tabular}
\end{table}

\subsection{Data Source and Evaluation Split}

CST-Bench curates scenes from the VG150 split of Visual Genome with valid
object boxes, sufficient distinct objects, and geometrically unambiguous
relations. The source annotations supply object categories, bounding boxes, and
scene relations used to identify valid spatial configurations. Pairwise
operators require two usable objects;
Third-object Relation Update and multi-object order operators require three
distinct object categories.

The benchmark contains 7,082 image-grounded samples across six intervention
families and twelve transition operators. The balanced evaluation split contains
480 examples, with 40 examples for each operator. Samples are partitioned by
source image so that source images in the evaluation split are excluded from
post-training data. This grouped split prevents a model from observing another
counterfactual variant of an evaluation image during training.

\begin{table*}[t]
\centering
\caption{Sample counts by transition operator. The official evaluation split
contains 40 examples per operator.}
\label{tab:operator_counts}
\small
\setlength{\tabcolsep}{4.5pt}
\begin{tabular}{lr|lr}
\toprule
\textbf{Transition operator} & \textbf{Samples} &
\textbf{Transition operator} & \textbf{Samples} \\
\midrule
Pairwise Left--Right Flip & 1,300 &
Lateral Relation Crossing & 386 \\
Multi-object Horizontal Order Flip & 300 &
Above--Below Crossing & 614 \\
Pairwise Relation Swap & 300 &
Vertical Near--Far Crossing & 300 \\
Third-object Relation Update & 1,300 &
Depth Order Change & 300 \\
Metric Distance Change (Relation Preserved) & 1,300 &
Front-to-Back Viewpoint Shift & 382 \\
Distance-to-Reference Near--Far Crossing & 300 &
Side-to-Side Viewpoint Shift & 300 \\
\midrule
\multicolumn{3}{r}{\textbf{Total}} & \textbf{7,082} \\
\bottomrule
\end{tabular}
\end{table*}

\subsection{Spatial Relation Definitions}

For a box $b_i=[x_i^1,y_i^1,x_i^2,y_i^2]$ in an image of width $W$ and height
$H$, the normalized center is
\begin{equation}
(\tilde{x}_i,\tilde{y}_i)=
\left(\frac{x_i^1+x_i^2}{2W},
      \frac{y_i^1+y_i^2}{2H}\right).
\end{equation}
For an ordered pair $(i,j)$, \textit{left of}/\textit{right of} are determined
by the sign of $\tilde{x}_i-\tilde{x}_j$, and \textit{above}/\textit{below}
are determined by the sign of $\tilde{y}_i-\tilde{y}_j$. Candidates with
insufficient center separation along the queried axis are rejected.

Normalized center distance is
\begin{equation}
d(i,j)=\frac{\lVert c_i-c_j\rVert_2}{\sqrt{W^2+H^2}}.
\end{equation}
Near--far operators retain only pairs outside an ambiguity band around the
qualitative boundary. A relation-preserved sample changes $d(i,j)$ without
changing its qualitative label, whereas a threshold-crossing sample changes
both the metric distance and the qualitative relation.

All boxes must have positive extent and remain within the image boundary before
and after the symbolic edit. Thus, every object needed to answer a retained
question remains visible. Front--behind labels are taken from available scene
annotations or from the operator-defined symbolic update; they are not inferred
from two-dimensional box centers.

\subsection{Complete Operator Specification}

Table~\ref{tab:operator_spec} defines the twelve operators. ``Affected'' lists
the relation updated by the symbolic editor; ``invariants'' lists the properties
that must remain unchanged for a valid transition.

\begin{table*}[t]
\centering
\caption{Complete CST-Bench operator specification.}
\label{tab:operator_spec}
\scriptsize
\setlength{\tabcolsep}{3pt}
\begin{tabularx}{\textwidth}{p{2.0cm}p{2.8cm}YYYY}
\toprule
\textbf{Family} & \textbf{Operator} & \textbf{Precondition} &
\textbf{Intervention} & \textbf{Affected / expected} &
\textbf{Principal invariants} \\
\midrule
Global mirror & Pairwise Left--Right Flip & Clear horizontal pair &
Mirror image coordinates & left $\leftrightarrow$ right & identity, count,
vertical relation \\
Global mirror & Multi-object Horizontal Order Flip & Three ordered, distinct
objects & Mirror all selected objects & reverse all horizontal pair orders &
identity, count, vertical relations \\
Object swap & Pairwise Relation Swap & Two distinct objects & Exchange their
positions & queried-pair relation reverses & identity, count, attributes \\
Object swap & Third-object Relation Update & Swapped pair plus unchanged
reference & Exchange two positions & update relation to third object & reference
object, identity, count \\
Metric distance & Metric Distance Change (Relation Preserved) & Clear near/far
pair & Change subject--reference distance & metric distance changes; label
preserved & direction, identity, image-plane order \\
Metric distance & Distance-to-Reference Near--Far Crossing & Clear pair with
valid offset & Change distance without crossing reference object & near
$\leftrightarrow$ far & side of reference, identity, orthogonal relation \\
Translation & Lateral Relation Crossing & Clear horizontal pair & Translate
subject across reference & left $\leftrightarrow$ right & identity, count,
vertical relation \\
Translation & Above--Below Crossing & Clear vertical pair & Translate subject
across reference & above $\leftrightarrow$ below & identity, count,
horizontal relation \\
Translation & Vertical Near--Far Crossing & Clear pair with valid vertical
offset & Translate vertically without crossing reference & near
$\leftrightarrow$ far & identity, count, horizontal relation \\
Depth & Depth Order Change & Source front/behind label & Move subject across
reference along depth & front $\leftrightarrow$ behind & identity, count,
left/right, above/below \\
Viewpoint & Front-to-Back Viewpoint Shift & Observer-centered depth relation &
Reverse viewing direction & front $\leftrightarrow$ behind & physical scene and
allocentric layout \\
Viewpoint & Side-to-Side Viewpoint Shift & Observer-centered lateral relation &
View scene from opposite side & left $\leftrightarrow$ right & physical scene
and allocentric layout \\
\bottomrule
\end{tabularx}
\end{table*}

\subsection{Constraint-Aware Candidate Generation}

The principal construction difficulty is that the post-intervention state is
not directly visible in the source image. CST-Bench addresses this issue with a
deterministic, operator-aware editor and symbolic solver. Starting from a
curated source state, the editor applies an intervention and the solver jointly
derives the source state, target state, affected relations, expected effects,
invariants, applicable metric changes, and final answer. Generating these
fields from one edited state prevents contradictions between a transition and
its dense annotations.

Candidates are discarded if they violate an operator-specific constraint,
contain invalid boxes or ambiguous references, lack sufficient relation
separation, fail to realize the intended effect, or move a referenced object
outside the image. Remaining candidates are scored by
\begin{equation}
s(c)=s_{\mathrm{ref}}+s_{\mathrm{rel}}+s_{\mathrm{edit}}
+s_{\mathrm{vis}}+s_{\mathrm{nat}},
\end{equation}
where the five terms respectively measure reference clarity, relation clarity,
edit effectiveness, post-edit visibility, and transformation naturalness.
Candidates with $s(c)<8$ are removed. The remaining candidates are ranked by
\begin{equation}
r(c)=\frac{1}{2}\hat{s}_{\mathrm{op}}(c)
+\frac{1}{2}\hat{s}_{\mathrm{global}}(c),
\end{equation}
where the two normalized scores are computed within an operator and globally,
respectively. At most one highest-ranked transformation is retained for each
source image.

\subsection{Answer-Blind Consensus Validation and Human Verification}

The symbolic label can be internally consistent while the image, reference, or
question remains ambiguous. CST-Bench therefore uses answer-blind consensus
validation before human verification. For each candidate, a VLM independently
solves the question $K=3$ times without seeing the symbolic answer. It then
compares its independent prediction with the symbolic label and scores five
criteria from 0 to 2: scene--image consistency, reference clarity,
transformation fidelity, answer correctness, and option quality.

A candidate is retained only when all three reviews recover the symbolic
answer, report no ambiguity, and assign the maximum score to every criterion:
\begin{equation}
\operatorname{Keep}(c)=
\bigwedge_{k=1}^{K}
\left[
\hat a_k=a^* \land b_k=0 \land \min_j q_{k,j}=2
\right],
\end{equation}
where $\hat a_k$ is the independent prediction, $a^*$ is the symbolic answer,
$b_k$ is the ambiguity indicator, and $q_{k,j}$ is the score of criterion $j$.
Candidates with repeated hard failures are excluded. Candidates without
unanimous agreement are routed to human verification rather than retained by a
majority vote.

Human verification follows a two-part protocol. First, two reviewers inspect
each unresolved candidate independently. They assess reference clarity,
intervention plausibility, source-to-target transition consistency, target
answer correctness, and distractor quality. A reviewer marks a sample as
\emph{retain}, \emph{correct}, or \emph{exclude}, with a brief rationale for
every correction or exclusion. Second, a stratified audit samples retained
examples from every operator to check that automated filtering does not conceal
operator-specific errors. When the two reviewers disagree, a third reviewer
adjudicates using the source image, the symbolic transition record, and the
operator definition in Table~\ref{tab:operator_spec}. Only adjudicated retain
or corrected decisions enter the final benchmark; ambiguous or inconsistent
samples are excluded.

\subsection{Released Data Fields}

Table~\ref{tab:data_fields} lists the principal fields in each released sample.
Structured fields can be distributed as JSON or as serialized values in a CSV
representation.

\begin{table}[H]
\centering
\caption{Principal CST-Bench fields.}
\label{tab:data_fields}
\scriptsize
\begin{tabularx}{\columnwidth}{p{2.5cm}Y}
\toprule
\textbf{Field} & \textbf{Meaning} \\
\midrule
\texttt{sample\_uid} & Stable sample identifier. \\
\texttt{image\_id} & Source-image identifier used for grouped splitting. \\
\texttt{operator} & Intervention family and fine-grained transition operator. \\
\texttt{question} & Intervention-conditioned natural-language query. \\
\texttt{options}, \texttt{answer} & Answer choices and target choice. \\
\texttt{source\_state} & Relevant source objects and spatial relations. \\
\texttt{target\_state} & Symbolically updated post-intervention state. \\
\texttt{intervention} & Operator and its object-specific parameters. \\
\texttt{affected\_relations} & Relations changed by the intervention. \\
\texttt{expected\_effects} & Required pre-to-post relational or metric changes. \\
\texttt{invariants} & Properties that must remain unchanged. \\
\texttt{metric\_change} & Metric-distance information, when applicable. \\
\bottomrule
\end{tabularx}
\end{table}

\FloatBarrier

\section{CT-OPSD Implementation Details}
\label{app:ctopsd}

\subsection{Teacher Supervision and Student Inputs}

Let $x=(I,q)$ denote the observable image-question input. For training, each
CST-Bench example additionally provides the causal transition annotations
\begin{equation}
z_c=(s^0,u,e,\mathcal{I},s^1),
\end{equation}
where $s^0$ and $s^1$ are the source and target states, $u$ is the
intervention, $e$ records affected relations and expected effects, and
$\mathcal{I}$ is the set of preserved invariants. Metric information is
included in $e$ when applicable. Together with the answer $a^*$, these labels
define the reference transition
\begin{equation}
\tau^*:s^0\xrightarrow{\,u\,}(e,\mathcal{I})
\xrightarrow{}s^1\xrightarrow{}a^*.
\end{equation}

The causal teacher is supervised to generate this structured transition from
$x$. Its teacher-training loss is
\begin{equation}
\mathcal{L}_{\mathrm{teacher}}
=-\sum_{t=1}^{|\tau^*|}
\log\pi_\phi(\tau_t^*\mid x,\tau^*_{<t}).
\end{equation}
The privileged labels specify the teacher target; they are not supplied as
inputs to the student. The teacher is frozen before student training.

The student is initialized from the same base model and receives only $x=(I,q)$
throughout rollout training and inference. It generates an on-policy trajectory
$\hat\tau\sim\pi_\theta(\cdot\mid x)$. At step $t$, the frozen teacher and
student evaluate the same student-generated prefix:
\begin{equation}
p_T^{(t)}=\pi_\phi(\cdot\mid x,\hat\tau_{<t}),\qquad
p_S^{(t)}=\pi_\theta(\cdot\mid x,\hat\tau_{<t}).
\end{equation}
The resulting causal bridge transfers teacher behavior at states visited by the
student rather than only along reference trajectories.

\subsection{Training Objective and Inference}

CT-OPSD minimizes forward KL divergence along student rollouts together with
standard answer supervision:
\begin{equation}
\mathcal{L}_{\mathrm{OPSD}}
=\mathbb{E}_{\hat\tau\sim\pi_\theta(\cdot\mid x)}
\left[
\frac{1}{L}\sum_{t=1}^{L}
D_{\mathrm{KL}}\!\left(p_T^{(t)}\parallel p_S^{(t)}\right)
\right],
\end{equation}
\begin{equation}
\mathcal{L}_{\mathrm{CT\text{-}OPSD}}
=\mathcal{L}_{\mathrm{OPSD}}
+\lambda_{\mathrm{ans}}
\left[-\log\pi_\theta(a^*\mid x)\right].
\end{equation}
At inference, the teacher and all privileged causal-transition annotations are
discarded. The retained student predicts from the image and question alone.

\subsection{Prompt and Output Format}

The student uses the same answer-only input as the evaluated foundation models:
the image, intervention-conditioned question, and answer options. It is not
given source-state labels, affected relations, expected effects, invariants,
target states, or metric annotations. The teacher is trained to emit a compact
causal transition chain followed by the answer; the student is distilled on
its own generated continuations and is evaluated by its final multiple-choice
answer.

\begin{table}[t]
\centering
\caption{CT-OPSD components and their roles.}
\label{tab:ct_components}
\small
\begin{tabularx}{\columnwidth}{p{2.5cm}Y}
\toprule
\textbf{Component} & \textbf{Role} \\
\midrule
Observable input $x$ & Image, question, intervention description, and answer options. \\
Causal teacher & Predicts source state, intervention consequences, invariants, target state, and answer. \\
Student rollout & Samples a trajectory from the observation-only student policy. \\
On-policy alignment & Matches the teacher and student distributions on the shared student prefix. \\
Answer supervision & Anchors the final response to the gold answer $a^*$. \\
Inference student & Uses only the image and question after teacher removal. \\
\bottomrule
\end{tabularx}
\end{table}

\subsection{Ablation Definitions}

The component ablations in the main paper remove one source of causal
supervision while keeping the base model, observable input, evaluation split,
and answer format fixed. \textbf{CT w/o Aux.} removes auxiliary structured
supervision beyond the final answer. \textbf{CT w/o Bridge} removes the
causal-transition bridge between teacher and student rollouts. \textbf{CT w/o
Eff.}, \textbf{CT w/o Inv.}, and \textbf{CT w/o Metric} remove the
corresponding effect, invariant, and metric information from the teacher's
causal-transition supervision. Standard \textbf{OPSD} does not use
CST-Bench's structured causal-transition annotations. All variants are
evaluated observation-only, using the image and question without privileged
annotations.

\section{Evaluation Protocol and Additional Results}
\label{app:evaluation}

\subsection{Answer-Only Evaluation}

At inference, every foundation model and post-trained student receives the
source image, counterfactual question, and answer options. Models do not access
source states, intervention effects, invariant relations, target states, or
other privileged annotations. Decoding uses a common answer-only prompt and
the response is parsed as a multiple-choice answer; invalid or unmatched
responses receive zero credit.

Let $a_i\in\{0,1\}$ indicate whether prediction $i$ is correct. For family
$f$ with index set $D_f$,
\begin{equation}
\operatorname{Acc}_f=\frac{1}{|D_f|}\sum_{i\in D_f}a_i,
\qquad
\operatorname{Macro}=\frac{1}{6}\sum_{f=1}^{6}\operatorname{Acc}_f,
\end{equation}
and
\begin{equation}
\operatorname{Micro}=\frac{1}{480}\sum_{i=1}^{480}a_i.
\end{equation}
Macro Avg. weights each family equally, whereas Micro Acc. weights every
example in the balanced 480-example evaluation split equally.

\subsection{Causal-Transition Diagnostics}

The diagnostic evaluation measures source-state understanding (Src.), effect
reasoning (Eff.), invariance identification (Inv.), and transition consistency
(Cons.). Source-state understanding assesses whether the model recovers the
relevant pre-intervention relation. Effect reasoning assesses whether it
identifies the intervention-induced change. Invariance identification checks
whether unaffected properties are preserved, and transition consistency checks
whether the predicted relation update agrees with the final answer.

For ablations, CST W-Avg. aggregates answer accuracy and the structured
diagnostics according to the predefined benchmark weights:
\begin{equation}
\mathrm{CST\ W\text{-}Avg.}=
\frac{\sum_m w_m S_m}{\sum_m w_m},
\end{equation}
where $S_m$ is the score of diagnostic dimension $m$ and $w_m$ is its fixed
weight. This aggregate is distinct from family-level Macro Avg. and is used to
compare whether a method jointly captures source states, effects, invariants,
and consistent target transitions.

\subsection{External Spatial Benchmarks}

Post-trained variants are also evaluated on 3DSRBench, CV-Bench, and the
spatial-relation and relative-depth subsets of BLINK$_{\mathrm{val}}$. All
post-training methods use the same base model and the same evaluation protocol
within each external benchmark. CST-Bench metrics are not combined with these
external benchmark scores; the reported external columns retain the standard
task-level accuracy or average for each benchmark.

\section{Limitations, Ethics, and Release Scope}
\label{app:limitations}

\paragraph{Geometric approximation.}
Bounding boxes support controlled image-plane relations and normalized distance
changes, but they are not metric three-dimensional reconstructions. Apparent
distance depends on projection, object scale, and depth. Depth transitions
therefore rely on source annotations and symbolic state updates rather than
two-dimensional box geometry alone.

\paragraph{Symbolic interventions.}
The observed image is real, but the post-intervention state is symbolic rather
than a photorealistically rendered edit. CST-Bench tests state updating under a
hypothetical intervention; it does not evaluate image-edit realism, physical
simulation, collision, deformation, or long-horizon dynamics.

\paragraph{Validation limits.}
Answer-blind consensus validation and human verification reduce ambiguity but
cannot remove all systematic errors. VLM reviewers can share model-specific
biases, and human judgments can remain uncertain for borderline spatial cases.
The released validation records should therefore be retained and extended when
the benchmark is revised.

\paragraph{Data and task coverage.}
The benchmark is English-language, multiple-choice, and limited to twelve
operators. Its visual, object, language, geographic, cultural, and demographic
coverage inherits that of its source data. Performance may not transfer to
open-ended responses, videos, robotics, safety-critical systems, or physical
settings with richer dynamics.

\paragraph{Responsible use.}
The source images may depict people and everyday environments. CST-Bench does
not add identity labels or infer sensitive attributes. A high benchmark score
does not establish that a model is safe for autonomous navigation or robotic
control, where an incorrect spatial update can cause unsafe actions.

\end{document}